# Face Liveness Detection Based on Client Identity Using Siamese Network

Huiling Hao, Mingtao Pei

Beijing Laboratory of Intelligent Information Technology, Beijing Institute of Technology,

Beijing 100081, P.R. China

*Abstract*—Face liveness detection is an essential prerequisite for face recognition applications. Previous face liveness detection methods usually train a binary classifier to differentiate between a fake face and a real face before face recognition. The client identity information is not utilized in previous face liveness detection methods. However, in practical face recognition applications, face spoofing attacks are always aimed at a specific client, and the client identity information can provide useful clues for face liveness detection. In this paper, we propose a face liveness detection method based on the client identity using Siamese network. We detect face liveness after face recognition instead of before face recognition, that is, we detect face liveness with the client identity information. We train a Siamese network with image pairs. Each image pair consists of two real face images or one real and one fake face images. The face images in each pair come from a same client. Given a test face image, the face image is firstly recognized by face recognition system, then the real face image of the identified client is retrieved to help the face liveness detection. Experiment results demonstrate the effectiveness of our method.

## I. Introduction

With the increasing deployment of face recognition in many applications such as intelligent entrance guard system, security surveillance and intelligent human machine interface, its security concern becomes increasingly important. Many face liveness detection methods are proposed [1-9].

Most previous face liveness detection methods train a binary classifier to differentiate between a fake face and a real face before face recognition. The client identity information is not utilized in previous face liveness detection methods. However, in practical face recognition applications, the real face images of the clients are available to the face recognition system, and face spoofing attacks are always aimed at a certain client. Therefore, the client identity information can provide useful clues for face liveness detection.

In this paper, we propose a face liveness detection method based on the client identity using Siamese network. We detect face liveness after face recognition instead of before face recognition, that is, we detect face liveness with the client identity information. In training stage, we collect face image pairs to train a Siamese network. Each image pair consists of two face images. The two face images can be a real face image and a fake face image, or two real face images. The two face images in each pair come from a same client. The trained Siamese network can classify the input image pair as "two real" or "one fake one real". In testing stage, the input test face image is first identified by a face recognizer and the identity information of the test face image is obtained. Then the real face image of the identified client is retrieved. The retrieved real face image and the test face image are classified by the trained Siamese network. If the Siamese network classify these two images as "two real", then the input test face image is a real face image, otherwise, it is a fake image.

The rest of this paper is organized as follows: Section II describes related works on face liveness detection. Section III demonstrates the details of our method. Section IV shows the experimental results. And section V concludes this paper.

## II. Related Work

Existing face liveness detection methods can be categorized into three groups with respect to the clues used for liveness detection: motion-based methods, texture-based methods, and 3D shape-based methods.

Motion-based methods: Motion-based methods are mainly based on the fact that living face is dynamic. Given an image sequence, these methods attempt to capture facial response like eye blinking, mouth movement, and head pose, then exploit spatial and temporal features. Pan [6] et al proposed a real-time face liveness detection method using an ordinary webcam by recognizing spontaneous eye-blinks. In this method, they constructed different stages of blinking action, and then used these as criterion to determine whether the eyes are open or closed. Bao [7] et al proposed to detect face liveness based on the difference between the optical flow information real face and fake face. Singh et al. [23] distinguished fake faces from the real ones by detecting eye and mouth movements based on a haar classifier.

Texture-based methods: Texture-based methods use the texture distortions which are caused by secondary imaging to detect face liveness. On the basis of differences between live face image and forged face image in spectral composition, Li [4] et al proposed a face living detection method using two-dimensional Fourier transform. In this method, they transformed face images into frequency domain to complete the classification. Dhrubajyoti.D [5] et al distinguished the authenticity of face by analyzing the energy values in frequency domain. If the energy value of measured face is lower than the pre-set threshold, it's determined to be a fake face. Alotaibi et al. [24] employed the nonlinear diffusion to enhance edges in the input image, then used convolution neural networks to detect face liveness based on the enhanced edges.

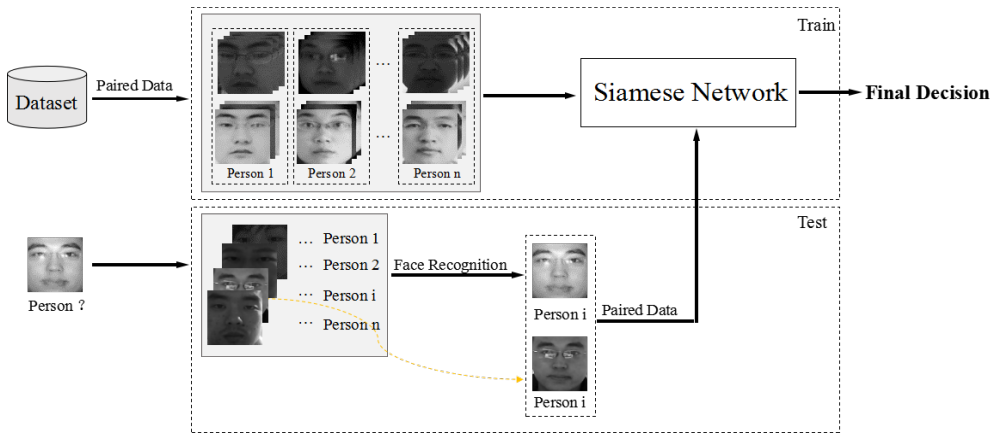

Fig. 1. The Framework of our method

3D shape-based methods: These methods are based on the fact that real face is 3-dimensional while fake face is usually 2-dimensional. However, these methods will fail when coping with 3D mask attacking, such as the 3D Mask Attack dataset (3DMAD) [25].

## III. PROPOSED METHOD

### A. Framework

In face spoof attacking, attackers may spoof the face recognitions system with photo, video or even 3D mask. Whatever the spoofing method, the only goal of the attacker is to let the face recognition system believe that the attacker is a certain client of the system. Usually, the face recognition system does have the real face image of its clients.

Based on the above observation, we propose a face liveness detection method based on the client identity. Fig 1 shows the framework of our method which contains two stages: offline training stage and online testing stage.

In the offline training stage, we collect face image pairs to train a Siamese network for the face liveness detection. Each image pair consists of two face images. The two face images can be a real face image and a fake face image, or two real face images. The two face images in each pair come from a same client. For an image pair, if its two face images are both real, it is a positive pair, otherwise, it is a negative pair. Fig. 2 shows two samples of the constructed positive and negative pairs. A Siamese network is trained on the positive and negative pairs. The trained Siamese network can classify the input image pair as "two real" or "one fake one real".

In testing stage, the input test face image is first identified by a face recognizer and the identity information of the test face image is obtained. Then the real face image of the identified client is retrieved. The retrieved real face image and the test face image form an image pair and are classified by the trained Siamese network. If the Siamese network classify these two images as "two real", then the input test face image is a real face image, otherwise, it is a fake image.

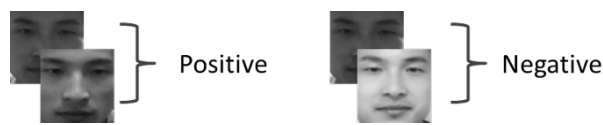

Fig. 2. Samples of constructed positive and negative pairs

### B. Siamese Network

To utilize the client identity information for face liveness detection, we use the Siamese network, which is proposed in [10] and modified for face verification in [11] [13]. Siamese Network is a class of neural network architectures that contain two or more subnetworks. The contained subnetworks may be identical or different. We use the Siamese network with two identical subnetworks. Fig. 3 shows the architecture of the Siamese network. The two subnetworks are convolution networks. They have the same configuration with the same parameters and weights, and parameter updating is mirrored across both subnetworks.

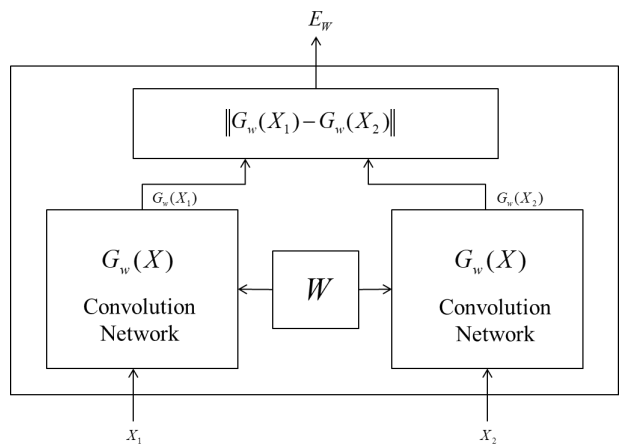

Fig. 3. *Siamese Architecture* [11]

The input of the Siamese network is an image pair $(X_1, X_2)$. Then image features are extracted by the two identical convolution neural network as $G_w(X_1)$ and $G_w(X_2)$, respectively. The contrastive loss function [26] is employed to train the network.

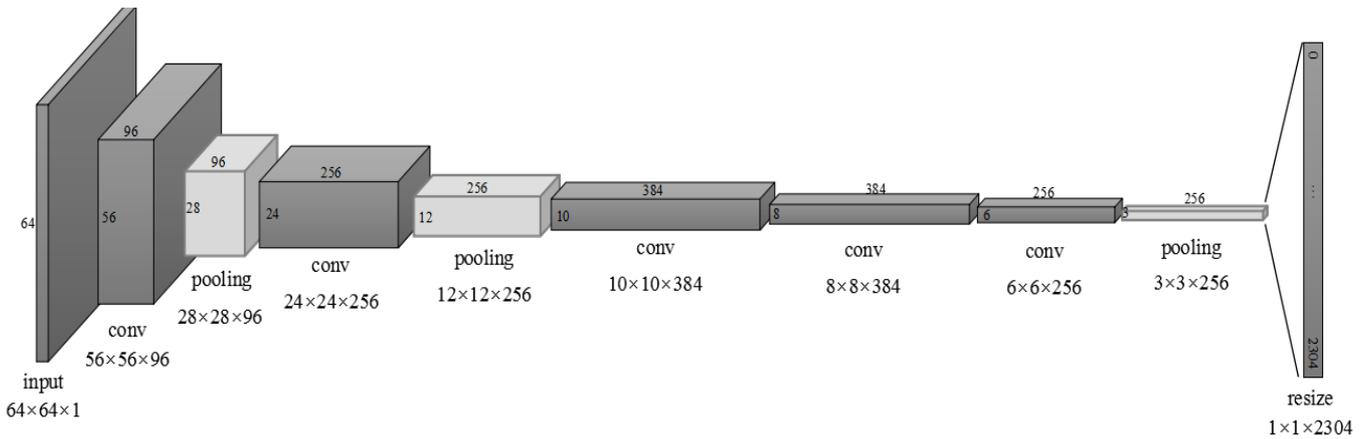

Fig. 4. The details of our convolution neural network

$$L = \frac{1}{2N}\sum_{n=1}^{N}(yd^2 + (1-y)\max(margin - d, 0)^2) \quad (1)$$

Where $d = \|a_n - b_n\|_2$ is the Euclidean distance between two samples' features, y represents the label. In our case, $y = 1$ indicates that the two face images are both real face images. $y = 0$ means that one of the two face images is fake face image. *Margin* is the pre-set threshold.

This loss function encourages matching pairs (two real face images of a person) to be close together in feature space while pushing non-matching pairs (one real face image and one fake face image of a person) apart.

In our implementation, the two subnetworks are based on the wildly used AlexNet [12] architecture. We make some modifications on the AlexNet to fit our data. The details of the convolution neural network we used are shown in Fig. 4. There are five convolution layers and three pooling layers. We use it to extract hierarchical features through multi-layers' convolution and obtain invariance property through pooling layers' down-sampling operation.

## IV. EXPERIMENTS

### A. Datasets

To prove the effectiveness of our method, we conduct experiments on two public datasets: NUAA [14] and Replay-Attack [15].

NUAA is a publicly available dataset, which is provided by Nanjing University of Aeronautics and Astronautics, and is widely used for the evaluation of face liveness detection. The data set contains 12614 images of 15 different subjects, including both real and fake face images. The database is divided into a training set with a total of 3491(real: 1,743 / fake: 1,748) images and a test set with a total of 9123 (real: 3,362 / fake: 5,761) images.

Replay-Attack is provided by IDIAP in 2012. It contains 1300 video clips of 50 different subjects. These video clips are divided into 300 real-access videos and 1000 spoofing attack videos. The dataset takes into consideration the different lighting conditions used in spoofing attacks. The Replay-Attack database consists of training set, development set and testing set.

To keep consistent to previous works, we use the half total error rate (HTER) as the metric in our experiments. The HTER is half of the sum of the false rejection rate (FRR) and false acceptance rate (FAR):

$$\text{HTER} = \frac{FRR + FAR}{2} \quad (2)$$

### B. Results on NUAA

In NUAA, for each real face image in the training set, we randomly select a real face image of the same subject from the training set to form a positive pair, and randomly select a fake face image of the same subject to form a negative pair. In total, we construct 1743 positive pairs and 1743 negative pairs to train the Siamese network.

In testing, for each image in the test set, we assume that the identity of the face in the test image is known (suppose the identity is $p_i$), we select one real face image of $p_i$ from the training set, and form an image pair with the test image. The formed image pair is input to the trained Siamese network to justify whether the test image is real of fake. We compare our method with the LBP [16], LBP+Gabor+HOG [17] and LBP+Gabor+Pixel [18]. Table I shows the comparison results of liveness detection on the NUAA dataset. We can see that our method performs better than the compared methods.

TABLE I.　　RESULTS ON NUAA DATASET

| Method | HTER(%) |
|---|---|
| LBP only [16] | 5.45 |
| LBP+Gabor+HOG [17] | 3.95 |
| LBP+Gabor+Pixel [18] | 2.45 |
| **Our method** | **1.96** |

### C. Results on Replay-Attack

For Replay-Attack dataset, we regard each video clip as a sequence of images. Similar to the NUAA dataset, we construct 14338 positive pairs and 14338 negative pairs to train the Siamese network. We compare our method with several state-of-art methods. Table II shows the comparison results of liveness detection on the Replay-Attack dataset.

TABLE II. RESULT ON REPLAY-ATTACK DATASET

| Method | HTER(%) |
|---|---|
| Fine-tuned VGG-Face [19] | 4.30 |
| DPCNN [19] | 6.10 |
| Boulkenafet et al. [20] | 2.90 |
| Boulkenafet et al. [21] | 2.20 |
| Moire pattern [22] | 3.30 |
| Patch-based CNN [2] | 1.25 |
| Depth -based CNN [2] | 0.75 |
| Patch and depth CNN [2] | 0.72 |
| **Our method** | **0.86** |

We can see that our method achieves good performance on the Replay-Attack dataset. The depth-based CNN [2] and Patch and depth CNN [2] performs better than our method. It may because we don't use of the depth information in the face liveness detection.

## V. CONCLUSION

In this paper, we propose a face liveness detection method based on the client identity information using Siamese network. Different from most of previous methods, we do the face liveness detection after face recognition. Therefore, the client identity information and the real face image of the client can be used to help the face liveness detection. In future work, we will consider using client identity information for face liveness detection in videos.